\documentclass[10pt,twocolumn,letterpaper]{article}

\usepackage{cvpr}              

\usepackage{graphicx}
\usepackage{amsmath}
\usepackage{multirow}
\usepackage{amssymb}
\usepackage{booktabs}
\usepackage{float}
\usepackage{newtxmath}
\usepackage{color, xcolor}
\usepackage{soul}
\soulregister\cite7 
\soulregister\citep7
\soulregister\citet7 
\soulregister\ref7 
\soulregister\pageref7 

\usepackage{xcolor}

\newcommand\equalcontribution{\thanks{Equal contribution.}}
\newcommand\internship{\thanks{Work done when Zhichao and Leshan were interns at VIS, Baidu.}}

\usepackage[pagebackref,breaklinks,colorlinks]{hyperref}

\usepackage[capitalize]{cleveref}
\crefname{section}{Sec.}{Secs.}
\Crefname{section}{Section}{Sections}
\Crefname{table}{Table}{Tables}
\crefname{table}{Tab.}{Tabs.}

\def\cvprPaperID{10087} 
\def\confName{CVPR}
\def\confYear{2022}

\begin{document}

\title{Human-Object Interaction Detection via Disentangled Transformer}

\author{Desen Zhou$^{1}$\equalcontribution\hspace{0.3cm} Zhichao Liu$^{1,2}$\footnotemark[1]\hspace{0.15cm}\internship\hspace{0.3cm} Jian Wang$^{1}$\hspace{0.2cm} Leshan Wang$^{1,2}$\footnotemark[2]\hspace{0.3cm} Tao Hu$^1$\hspace{0.2cm} Errui Ding$^{1}$\hspace{0.2cm} Jingdong Wang$^{1}$\\
$^{1}$Department of Computer Vision Technology (VIS), Baidu Inc.\\
$^{2}$ShanghaiTech University\\
\tt\small \{zhoudesen,wangjian33,hutao06,dingerrui\}@baidu.com \\ 
\tt\small \{liuzhch,wanglsh\}@shanghaitech.edu.cn, wangjingdong@outlook.com}


\maketitle

\begin{abstract}

Human-Object Interaction Detection tackles the problem of joint localization and classification of human object interactions. Existing HOI transformers either adopt a single decoder for triplet prediction, or utilize two parallel decoders to detect individual objects and interactions separately, and compose triplets by a matching process. In contrast, we decouple the triplet prediction into human-object pair detection and interaction classification. Our main motivation is that detecting the human-object instances and classifying interactions accurately needs to learn representations that focus on different regions. To this end, we present Disentangled Transformer, where both encoder and decoder are disentangled to facilitate learning of two sub-tasks. To associate the predictions of disentangled decoders, we first generate a unified representation for HOI triplets with a base decoder, and then utilize it as input feature of each disentangled decoder. Extensive experiments show that our method outperforms prior work on two public HOI benchmarks by a sizeable margin. Code will be available.

\end{abstract}

\section{Introduction}
\label{sec:intro}

\begin{figure}[t]
\centerline{\includegraphics[width=0.5\textwidth]{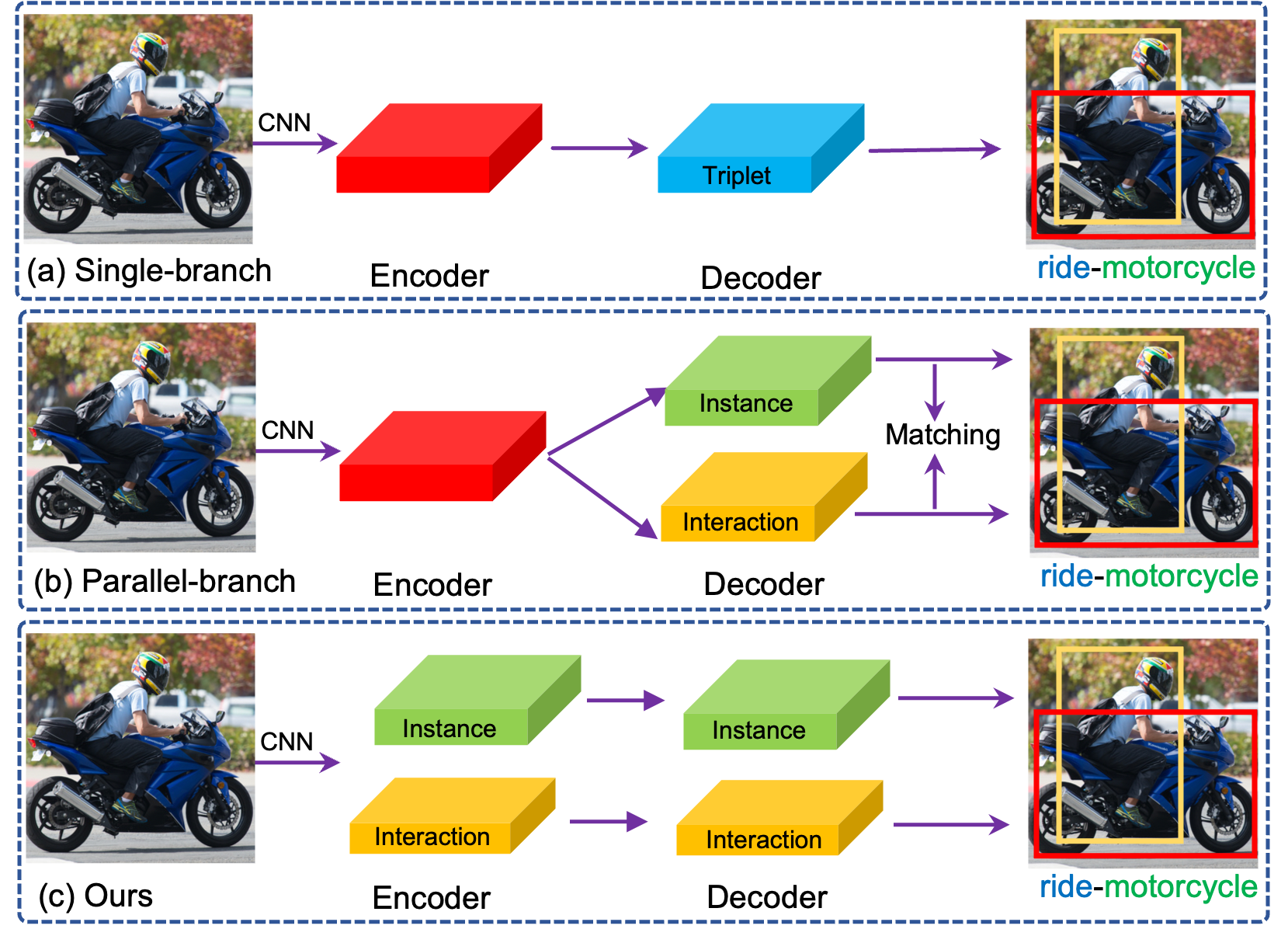}}
\caption{Architecture comparison of different HOI transformers. (a) Single-branch transformer\cite{qpic,hoitrans} adopts a single decoder to directly detect HOI triplets. (b) Parallel-branch transformer\cite{asnet,hotr} utilizes separate decoders detect individual objects and interactions, and then compose triples by a matching process, which might introduce additional grouping errors. (c) Ours disentangles the task of triplet prediction into human-object pair detection and interaction classification via an instance 
stream and an interaction stream, where both encoder and decoder are disentangled.}
\label{fig:adv}
\end{figure}

Human-object interaction(HOI) detection\cite{vcoco} aims at detecting all the $<$human, verb, object$>$ triplets in an image. It has attracted increasing attention in the computer vision community in recent years\cite{gao2018ican,gkioxari2018detecting}. Accurate estimation of human-object interactions can benefit multiple downstream tasks, such as human action recognition\cite{yan2018spatial}, scene graph generation\cite{lin2020gps}, and image caption\cite{chen2020say}.


\par


Recent advances show that HOI detection can be formulated as set prediction problem\cite{hotr,asnet,qpic,hoitrans}. Existing HOI transformers can be categorized into two types: single-branch transformer and parallel-branch transformer. Single-branch transformer\cite{qpic,hoitrans} adopts multi-task strategy, in which one query is responsable for predicting a $<$human, verb, object$>$ triplet within a single decoder. In contrast, parallel-branch transformer\cite{hotr,asnet} adopts parallel decoders for instance detection and interaction classification separately. Specifically, one instance decoder follows DETR\cite{detr} and detects individual objects, and the other interaction decoder estimates the interactions in the image. To compose HOI triplets, it generates additional associative embeddings to match the interactions and instances. Since HOI detection is a composition problem\cite{hou2021detecting,hou2020visual}, the latter decomposing strategy has several advantages compared with unified multi-tasking strategy. First, two sub-task decoders might attend to different regions via cross attention to facilitate learning and also results in better interpretability. In addition, it has better generalizability, especially for rare categories due to long-tail distribution of triplet compositions. However, existing parallel-decoder transformers suffer from two crucial drawbacks under complex scenarios: i) the interaction predictions have to find their corresponding human and object instances in instance decoder, which might introduce additional errors due to mis-grouping; ii) regardless of the shared encoder, the decoding sub-tasks are relatively independent and the joint configurations of instances and interactions are not considered.

To overcome above limitations, we present Disentangled Transformer(DisTR). We decouple the triplet prediction into human-object pair detection and interaction classification via an instance stream and an interaction stream, where both encoder and decoder are disentangled. An illustration of architecture comparison between ours and prior HOI transformers is shown in Fig.\ref{fig:adv}. Our encoder module extracts different contextual information for two sub-tasks. During decoding process, the task decoder decodes its representation based on the corresponding task encoder. Different from prior parallel-decoder transformers\cite{hotr,asnet} that the instance decoder predicts individual objects, our instance decoder predicts a set of interactive human-object pairs. To associate the predictions of task decoders, we adopt a base decoder to first generate a unified representation for HOI triplets, following QPIC\cite{qpic}, and then utilize it as input feature of each task decoder. The task decoder then refines its representation based on the unified representation, resulting in a coarse-to-fine process. We further design an attentional fusion block to pass information between task decoders help them communicate with each other.

We evaluate our proposed method on two public benchmarks: V-COCO\cite{vcoco} and HICO-DET\cite{hico}. Our method outperforms current state-of-the-art by a sizeable margin. We further visualize the cross attentions in our task decoders, and observe that our task decoders indeed attend to different spatial regions, demonstrating the effectiveness of our proposed disentangled strategy.

The contributions of this paper are three folds:
\begin{itemize}
    \item We propose a disentangled strategy for HOI detection, where the triplet prediction is decoupled into human-object pair detection and interaction classification via an instance stream and an interaction stream.
    \item We develop a new transformer, where both encoder and decoder are disentangled. We also propose a coarse-to-fine strategy to associate the predictions of instance decoder and interaction decoder, and an attentional fusion block for communication between task decoders.
    \item We achieve new state-of-the-art on both V-COCO and HICO-DET benchmarks.
\end{itemize}

\section{Related Work}

\subsection{Two-stage Methods}
A classical branch of research to HOI detection are based on the hypothesis-and-classify strategy, which first detects object instances via object detectors\cite{girshick2015fast,ren2015faster}, and then perform interaction classification on the grouped pairwise human-object proposals\cite{gao2018ican,gkioxari2018detecting,wan2019pose,li2019transferable,li2020pastanet,liu2020amplifying}. Some works also exploit graph structure to enhance object dependencies\cite{zhang2021spatially, qi2018learning, ulutan2020vsgnet, wang2020contextual, liu2020consnet}. Another bunch of two-stage methods is the compositional approaches\cite{li2020hoi,hou2021affordance,hou2020visual,hou2021detecting}, which disentangle HOI representations by learning from fabricated compositional HOIs. In contrast, our method disentangles representations by disentangled task encoders and decoders and its one-stage framework does not rely on pre-computed object proposals.



\begin{figure*}[tbp]
\centerline{\includegraphics[width=1.\textwidth]{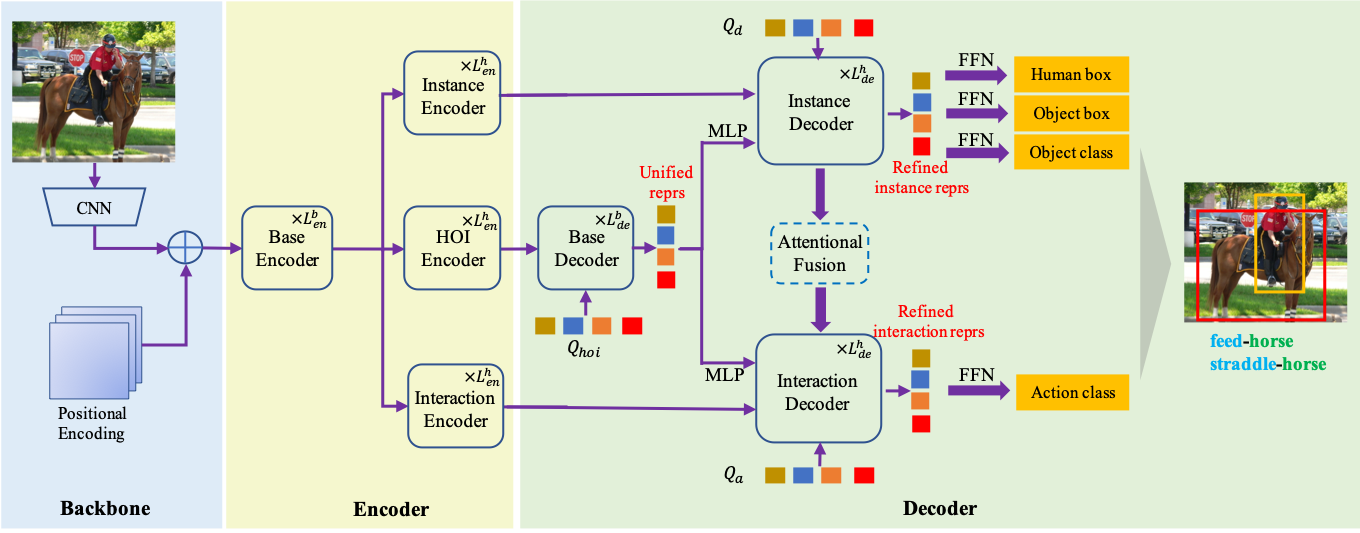}}
\caption{Overview of our framework. \textbf{Encoder module} extracts global contexts at three levels for different decoding sub-tasks. \textbf{Decoder module} disentangles the representations of instances and interactions in a coarse-to-fine manner: the \textit{base decoder} extracts unified HOI representation of HOI triplets, then the \textit{instance decoder} and \textit{interaction decoder} refines the unified HOI representation in disentangled feature spaces. Our instance decoder directly estimates interactive human-object instance pairs, which are associated with interaction predictions. The \textit{Attentional fusion blocks} are further inserted at each output layer(excluding the last layer) of two task decoders to perform communication between them.}
\label{overview}
\end{figure*}

\subsection{One-stage Methods}
Recently, one-stage or parallel HOI has caused extensive concern which transforms the interaction target as a center point or interaction object, and then adopt a detection pipeline. PPDM\cite{liao2020ppdm} which is based on CenterNet\cite{duan2019centernet} detects the interaction centers as well as objects, and then perform grouping as its post-process. IP-Net\cite{wang2020learning} is similar. UnionDet\cite{kim2020uniondet} use a novel union-level detector that eliminates this additional inference stage by directly capturing the region of interaction. DIRV\cite{fang2020dirv} concentrates on the densely sampled interaction regions across different scales for each human-object pair and introduce a novel voting strategy to replace Non-Maximal Suppression(NMS). 

\vspace{-0.3cm}
\paragraph{HOI Transformer} Recent HOI transformers follow DETR\cite{detr}, but separate into two types: entangled transformer and disentangled transformer. The entangled transformer, QPIC\cite{qpic} and HOITrans\cite{hoitrans} directly generate multiple $<$human,object,action$>$ triples of given image with a single decoder. On the contrary, disentangled transformers, HOTR\cite{hotr} and ASNet\cite{asnet} predict the objects and interactions in parallel decoders, and then perform matching between objects and interaction targets to generate final predictions. Recently, Zhang et.al\cite{zhang2021mining} propose to disentangle the instance decoder and interaction decoder in a cascaded process, which treats the instance decoder as proposal generator to interaction decoder. In contrast, our sub-tasks are parallelly decoded, hence the communication can be applied. In addition, our disentanglement is more complete due to encoder disentanglement.

\section{Method}
\subsection{Overview}
We adopt the one-stage transformer framework, which directly estimates all the $<$human,verb,object$>$ triplets given an image. To achieve this, we first group the HOI triplets with the same human and object instances. Then, the ground truth of an image can be represented as a tuple set $\{(\tilde{\mathbf{x}}_h^i, \tilde{\mathbf{x}}_o^i, \tilde{\mathbf{c}}^i, \tilde{\mathbf{a}}^i)|i=1,2,...,M\}$, where $M$ is the number of ground truth human-object interaction pairs, $
\tilde{\mathbf{x}}_h^i, \tilde{\mathbf{x}}_o^i\in\mathbb{R}^4$ denote the bounding boxes of human instance and object instance, $\tilde{\mathbf{c}}\in\{0,1\}^\mathcal{C}$ indicates the one-hot encoding of object category and $\mathcal{C}$ is the number of object classes, $\tilde{\mathbf{a}}^i\in\{0,1\}^{\mathcal{A}}$ denotes the labels of $\mathcal{A}$ interaction classes. We then deploy our transformer network to predict such tuple set. Formally, given image $I$, our goal is to define a transformer network $\mathcal{F}$ that performs the mapping:
\begin{equation}
    I \xrightarrow[]{\mathcal{F}}\{(\mathbf{x}_h^i, \mathbf{x}_o^i, \mathbf{c}^i, \mathbf{a}^i)|i=1,2,...,N_q\},
\end{equation}
\noindent where $i$ is the query index and $N_q$ is the number of queries pre-defined in our transformer network. $\mathbf{x}_h^i, \mathbf{x}_o^i\in \mathbb{R}^{4}$ denote the predicted bounding boxes of human instance and object instance respectively, $\mathbf{c}_o^i\in(0,1)^{\mathcal{C}+1}$ is estimated probability of object classification, which is normalized by $\mathrm{softmax}$ function. The additional dimension indicates background non-object class. $\mathbf{a}^i\in(0,1)^{\mathcal{A}}$ indicates the interaction probabilities, which are normalized by $\mathrm{sigmoid}$ function. 


We adopt a coarse-to-fine strategy to disentangle the instance detection and interaction classification, to resolve the matching problem between predictions. Specifically, we first generate a unified HOI representation
to represent the HOI triplets $\{(\mathbf{x}_h^i, \mathbf{x}_o^i, \mathbf{c}^i, \mathbf{a}^i)\}$, then an instance decoder is utilized to refine the representation in instance space and predict the human-object instance pairs, indicated by $\{(\mathbf{x}_h^i, \mathbf{x}_o^i, \mathbf{c}^i)\}$. And the interaction decoder is responsable for interaction disentanglement and prediction, indicated by $\{\mathbf{a}^i\}$. During inference, the predictions of the same query index in two head decoders are directly grouped together. Below we introduce our detailed implementation of above coarse-to-fine disentangling strategy.

\subsection{Network Architecture}
Similar to existing HOI transformers\cite{qpic,asnet} and DETR\cite{detr}, our network consists of three main modules: backbone module computes image features; encoder module exploits self-attention mechanism to further extract higher relational contexts between different spatial regions; and decoder module extracts representations from encoder module for the disentangled sub-tasks of instance detection and interaction classification. An overview of our framework is shown in Fig.\ref{overview}.
\subsubsection{Backbone module}
A CNN backbone is used to extract the high level semantic feature map with shape $(H, W, C)$, and then a $1 \times 1 $ convolution layer is used to reduce the channel dimension from $C$ to $D$. We flatten the feature map of shape $(H, W, D)$ to $(HW, D)$. We utilize ResNet50\cite{he2016deep} as our backbone, and reduce the feature map in conv-5 using $1\times 1$ convolution from $C=2048$ to $D=256$, the backbone visual features are represented as $\Gamma_{back} \in \mathbb{R}^{{HW}\times {D}}$.
\subsubsection{Encoder module}
Our encoder module aims at modeling relationships at different spatial regions to enhance global contexts for backbone representation $\Gamma_{back}$. Prior parallel-decoder transformers\cite{hotr,asnet} utilize shared encoder for instance detection and interaction classification. However, we assume that the relations in image representations of different sub-tasks are different and the encoder representations better be designed for specific sub-tasks. Hence we disentangle our encoder at three levels for different decoding sub-tasks: human-object pair detection, interaction classification and unified representation generation. Specifically, it consists of a base encoder and three head encoders. The base encoder, which consists of $L_{en}^{b}$ layers, enhances $\Gamma_{back}$ to generate a base encoder representation $\Gamma_{en}^{b}$. Then, three different head encoders with $L_{en}^{h}$ layers refine the base encoder representation separately. We denote the refined head representations as $\Gamma_{en}^{hoi}, \Gamma_{en}^{d}, \Gamma_{en}^{a}$, which are used for computing cross attentions in different decoders: $\Gamma_{en}^{hoi}$ for base decoder, $\Gamma_{en}^{d}$ for instance decoder and $\Gamma_{en}^{a}$ for interaction decoder. All the encoder representations share the same shape: $\Gamma_{en}^{b}, \Gamma_{en}^{hoi}, \Gamma_{en}^{d}, \Gamma_{en}^{a} \in \mathbb{R}^{HW\times D}$.\footnote{In this section, `b' is short for base, `d' indicates detection/instance, `a' indicates action/interaction, `h' indicates head.}

\subsubsection{Decoder module}
Our decoder module adopts attention mechanism to extract representations from encoder for sub-task decoding. We disentangle the representations of instances and interactions in a coarse-to-fine manner, which first utilizes a base decoder to generate a unified representation for a HOI triplet, and then exploits another two disentangled decoders to refine the unified representation in the spaces of instances and interactions. Different from previous transformers\cite{asnet,hotr} that the instance decoder predicts individual objects regardless of their interactiveness, our instance decoder estimates interactive human-object instance pairs associated with the interaction prediction. Hence it requires no additional matching process. To further help two task decoders communicate with each other, we propose an attentional fusion block to pass information between them. Below we describe the detailed structures of above components.

\paragraph{Base decoder}
Our base decoder has $L_{de}^b$ layers and generates unified HOI representations for the disentangled decoders to facilitate feature refinements and associate predictions. Formally, the base decoder $\mathcal{F}_{de}^{b}$ transforms a set of learnable HOI queries $Q_{hoi}\in \mathbb{R}^{N_q\times D}$ into a set of base HOI representations $\Gamma_{de}^{b}\in\mathbb{R}^{N_q\times D}$ from HOI encoder head:
\begin{equation}
    \Gamma_{de}^{b} = \mathcal{F}_{de}^{b}(\vmathbb{0},\Gamma_{en}^{hoi}, \mathbf{p}_{en}, Q_{hoi}),
\end{equation}
where the zero matrix $\vmathbb{0}=\{0\}^{N_q\times D}$ indicates the input feature of base decoder. $\mathbf{p}_{en}\in\mathbb{R}^{HW\times D}$ is the position embedding of the encoder representations.

\paragraph{Instance decoder}
Our instance decoder aims at refining the unified HOI representation $\Gamma_{de}^{b}$ to generate a disentangled representation for interactive human-object instance pairs. To achieve this, we utilize a MLP to embed the unified representation to generate input feature of instance decoder. Our instance decoder $\mathcal{F}_{de}^{d}$ has $L_{de}^h$ layers, and takes the input feature, together with a set of learnable instance queries $Q_{d}\in \mathbb{R}^{N_q\times d}$ to perform feature refinement. We found that inputting the unified representation as decoder feature is better than directly utilizing it as queries, because the disentangled decoders will have a powerful initial feature. The output of the instance decoder is a set of interactive human-object instance pairs:
\begin{equation}
\{(\mathbf{x}_h^i, \mathbf{x}_o^i, \mathbf{c}^i)\} = \mathcal{F}_{de}^{d}(\mathrm{MLP}(\Gamma_{de}^{b}), \Gamma_{en}^{d}, \mathbf{p}_{en}, Q_{d}).
\end{equation}


\paragraph{Interaction decoder}
Similar to the instance decoder, our $L_{de}^h$-layer interaction decoder refines the unified HOI representation to the disentangled interaction feature space and generate a set of interaction predictions:
\begin{equation}
\{\mathbf{a}^i\} = \mathcal{F}_{de}^{a}(\mathrm{MLP}(\Gamma_{de}^{b}), \Gamma_{en}^{a}, \mathbf{p}_{en}, Q_{a}),
\end{equation}
where $Q_{a}\in\mathbb{R}^{N_q\times D}$ indicates the query set, $\Gamma_{en}^{a}$ is the representation of interaction encoder. Similar to instance decoder, during decoding, the estimated interactions are associated with unified HOI representation, as well as the human-object pairs in instance decoder.

\begin{figure}[t]
\centerline{\includegraphics[width=0.5\textwidth]{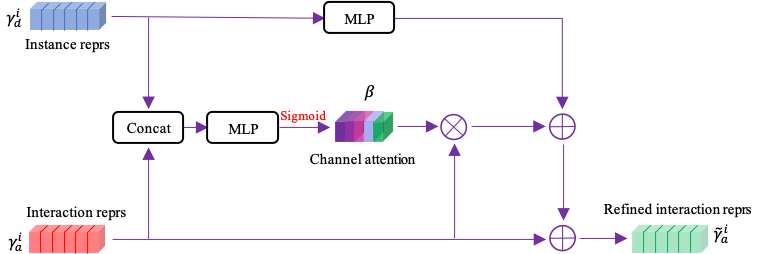}}
\caption{The structure of attentional fusion block.}
\label{fig:fusion}
\end{figure}

\paragraph{Attentional fusion block}
Our disentangled task decoders perform sub-tasks separately. However, two functional modules are not sufficiently communicated due to early decomposition of unified representations.\footnote{In our model, the instance decoder and interaction decoder have more layers than the base decoder.} To make the sub-tasks better benefit from each other, we perform message passing between the instance decoder and interaction decoder. Specifically, in the output of each layer in disentangled decoders, we fuse the instance representation to the interaction representation if they are associated with the same query index. The design of our fusion block is inspired by \cite{xiao2019reasoning} which utilizes the object representation and action representation to estimate a channel attention. Formally, we denote the instance representation and interaction representation for query $i$ as $\gamma_{d}^{i},\gamma_{a}^i\in\mathbb{R}^D$.  As shown in Fig.\ref{fig:fusion}, our attentional fusion block first concatenates the $\gamma_{d}^i$ and $\gamma_{a}^{i}$ and compute a channel attention $\beta\in \mathbb{R}^D$ with a MLP:
\begin{equation}
    \beta = \sigma(\mathrm{MLP}(\mathrm{Concat}([\gamma_{a}^i;\gamma_{d}^i]))),
\end{equation}
where $\sigma$ is the $\mathrm{sigmoid}$ function to constrain the elements in $\beta$ to range $(0,1)$. The channel attention is used to enhance the interaction representation with element-wise multiplication. During practice, we found that adding instance features provides additionally improvement. Hence, the output interaction representation $\tilde{\gamma}_{a}^i\in\mathbb{R}^D$ has the form:
\begin{equation}
    \tilde{\gamma}_{a}^i = \gamma_{a}^i + \beta \odot \gamma_{a}^i + \mathrm{MLP}(\gamma_{d}^i),
\end{equation}
where $\odot$ indicates the element-wise multiplication. In the last layer of disentangled decoders, we do not apply attentional fusion, in order to make the final representations more discriminative.

\subsection{Model Learning}
We adopt similar losses with previous HOI Transformer\cite{qpic,hoitrans}. Specifically, the instance decoder and interaction decoder generates set predictions $\{(\mathbf{x}_h^i,\mathbf{x}_o^i, \mathbf{c}^i)\}$ and $\{\mathbf{a}^i\}$, where $\mathbf{c}^i\in (0,1)^{\mathcal{C}+1}$, $\mathbf{a}^i\in (0,1)^{\mathcal{A}}$ indicate the object class probabilities and interaction class probabilities, which are normalized by $\mathrm{softmax}$ and $\mathrm{sigmoid}$ respectively. Then the predictions with the same query index are grouped together to a triplet set $\{(\mathbf{x}_h^i, \mathbf{x}_o^i, \mathbf{c}^i, \mathbf{a}^i)\}$. The rest process is the same as previous HOI transformers\cite{qpic} that first exploit the combined triplet predictions to compute a Hungarian Matching to the ground truth triplets, and then adopt different loss functions to the matched triplets. We denote $\mathcal{L}_b, \mathcal{L}_u, \mathcal{L}_c, \mathcal{L}_a$ as bounding box $L1$ losses, GIoU loss, object classification loss and interaction classification loss, the overall loss given by:
\begin{equation}
    \mathcal{L} = \lambda_b \mathcal{L}_b + \lambda_u \mathcal{L}_u + \lambda_c \mathcal{L}_c + \lambda_a \mathcal{L}_a,
\end{equation}
where $\lambda_b, \lambda_u, \lambda_c, \lambda_a$ denote the weights to balance the different loss components.
\paragraph{Auxiliary loss}
Inspired by DETR\cite{detr}, we add prediction FFNs and adopt auxiliary losses to each decoder layer to extract better representations. Our base decoder decodes unified HOI representations and predicts HOI triplets. The disentangled decoders predict instances and interactions respectively. Since the representations in disentangled decoders are refined from unified representation, we adopt different prediction FFNs to the disentangled decoders and base decoder. While in the same decoder, FFN parameters are shared.

\subsection{Model Inference}
Given HOI prediction set $\{(\mathbf{x}_h^i, \mathbf{x}_o^i, \mathbf{c}^i, \mathbf{a}^i)\}$, where $\mathbf{c}^i\in (0,1)^{C+1}$, $\mathbf{a}^i\in (0,1)^{\mathcal{A}}$ denote the classification probabilities of object class and action classes, the predicted object class and its detection score is given by $\operatorname{argmax}_{k} \mathbf{c}_{k}^{i}$ and $\operatorname{max}_{k} \mathbf{c}_{k}^{i}$, the output HOI of $j$-th action in $i$-th query is given by $(\mathbf{x}_h^i, \mathbf{x}_o^i, \operatorname{argmax}_{k} \mathbf{c}_{k}^{i}, j)$ with a prediction score $\operatorname{max}_{k} \mathbf{c}_{k}^{i}\cdot \mathbf{a}^i_j$. Similar to prior work\cite{qpic}, we only keep a prediction if its confidence score is above a threshold.

\section{Experiments}


\subsection{Experimental Setup}
\vspace{-0.1cm}
\paragraph{Dataset} We conducted experiments on two HOI detection datasets: HICO-DET\cite{hico} and V-COCO\cite{vcoco}. V-COCO is derived from MS-COCO\cite{lin2014microsoft} and contains 5400 and 4946 images in trainval subset and test subset respectively. V-COCO is annotated with 80 object categories and 29 action classes including 25 HOI triplets and 4 human body actions. HICO-DET contains 38118 and 9658 images for training and testing respectively. The HICO-DET has 80 object categories which is same as MS-COCO and 117 verb categories, all objects and verbs consist of 600 HOI triplets.
\vspace{-0.6cm}
\paragraph{Evaluation Metrics} Following prior work\cite{asnet,qpic,gao2018ican}, we use mean average precision(mAP). A triplets prediction is considered positive if human and object boxes have a IOU larger than 0.5 with ground truth boxes, and the predicted object categories and verb categories need to be correct. For HICO-DET, we report mAP over Full, Rare, and Non-Rare settings. For V-COCO, we report mAP on scenario \#1 (including objects) and scenario \#2 (ignore objects).



\subsection{Implementation Details}
In our implementation, the layer number of base encoder, head encoder, base decoder and head decoder are set to $L_{en}^b=4$, $L_{en}^h=2$, $L_{de}^b=2$, $L_{de}^h=4$. Query number $N_q=100$. We set the weight coefficients of $\lambda_b$, $\lambda_u$, $\lambda_c$, $\lambda_a$ to 2.5, 1, 1, 1. During training, we initialize our model parameters with pre-trained DETR\cite{detr} on COCO dataset. For the missing parameters, we adopt a warmup strategy, which first freezes the pre-trained parameters and adjusts the missing parameters for 10 epochs. Following prior work\cite{asnet,qpic}, we set the parameters in encoder and decoder to $10^{-4}$, and the backbone to $10^{-5}$. Weight decay is set to $10^{-4}$. Batch size is set to 16. For V-COCO, we freeze the backbone to avoid over-fitting. For HICO-DET, we fine-tune the whole model end-to-end. Including warmup, HICO-DET and V-COCO are trained with 80 epochs and learning rate is decreased at 65th epoch with 10 times. Our experiments are conducted on 8 Tesla V100 GPUs.

\subsection{Comparison to State-of-the-art}
We show the comparison of our method with previous two-stage and one-stage methods in Tab. \ref{V-COCO} and Tab. \ref{HICO-DET}. Our method outperforms prior works on both benchmarks. 

On V-COCO dataset, compared with state-of-the-art one-stage method QPIC\cite{qpic}, ours outperforms it with a significant gap. Compared with state-of-the-art two-stage method SCG\cite{zhang2021spatially}, our method also yields a large performance gap with 12.0\% mAP. It illustrates the our method has overwhelming advantage on both one-stage and two-stage methods. Particularly, our method outperforms previous parallel-branch HOI transformer HOTR\cite{hotr} and AS-Net\cite{asnet} by a large margin with 11.0\% mAP and 12.3\% mAP under scenario \#1.


\begin{table}[t]
    \centering
    \scalebox{0.85}{
    \begin{tabular}{c|c|cc}
    \hline
        Method & Backbone & Scenario \#1 & Scenario \#2 \\
        \hline
        Two-stage Method & & & \\
        iCAN\cite{gao2018ican} & R50 & 45.3 & 52.4 \\
        TIN\cite{li2019transferable} & R50 & 47.8 & 54.2 \\
        VCL\cite{hou2020visual} & R101 & 48.3 & - \\
        DRG\cite{gao2020drg} & R50-FPN & 51.0 & - \\
        VSGNet\cite{ulutan2020vsgnet} & R152 & 51.8 & 57.0 \\
        PMFNet\cite{wan2019pose} & R50-FPN & 52.0 & - \\
        PDNet\cite{zhong2020polysemy} & R152 & 52.6 & - \\
        CHGNet\cite{wang2020contextual} & R50 & 52.7 & - \\
        FCMNet\cite{liu2020amplifying} & R50 & 53.1 & - \\
        ACP\cite{kim2020detecting} & R152 & 53.2 & - \\
        IDN\cite{li2020hoi} & R50 & 53.3 & 60.3 \\
        SCG\cite{zhang2021spatially} & R50-FPN & 54.2 & 60.9 \\
        \hline
        One-stage Method & & & \\
        UnionDet\cite{kim2020uniondet} & R50-FPN & 47.5 & 56.2 \\
        IPNet\cite{wang2020learning} & HG104 & 51.0 & - \\ 
        GG-Net\cite{zhong2021glance} & HG104 &  54.7 & - \\
        DIRV\cite{fang2020dirv} &  EfficientDet-d3 & 56.1 & - \\
        HOITrans\cite{hoitrans} & R101 & 52.9 & - \\
        AS-Net\cite{asnet} & R50 & 53.9 & - \\
        HOTR\cite{hotr} & R50 & 55.2 & 64.4 \\
        QPIC\cite{qpic} & R50 & 58.8 & 61.0 \\
        Ours & R50 & \textbf{66.2} & \textbf{68.5}\\
        \hline
    \end{tabular}}
    \vspace{-0.1cm}
    \caption{\small Performance comparison on V-COCO test set.}
    \vspace{-0.1cm}
    \label{V-COCO}
\end{table}

\begin{table}[t]
    \centering
    \scalebox{0.9}{
    \begin{tabular}{c|cc}
    \hline
        Method &  Scenario \#1 & Default(Full) \\
        \hline
        Ours & \textbf{66.2} & \textbf{31.75} \\
        w/o encoder disentanglement & 65.5 & 30.79 \\
        w/o attentional fusion & 64.4 & 31.24 \\
        w/o decoder disentanglement & 58.8 & 29.07 \\
        \hline
    \end{tabular}}
    \vspace{-0.1cm}
    \caption{\small Ablation study of model components on both V-COCO test set (Scenario \#1) and HICO-DET test set (Default, Full setting)}
    \vspace{-0.5cm}
    \label{module ablation}
\end{table}

On HICO-DET dataset, compared with state-of-the-art one-stage methods, with R50 backbone, our method outperforms QPIC\cite{qpic} by 2.68\% mAP, and AS-Net\cite{asnet} by 2.88\% mAP under Default Full setting. It's also worth noting that under Rare setting, our method achieves 27.45\%, which is significant better than QPIC, demonstrating the effectiveness of disentangled strategy. Our method also outperforms recent state-of-the-art two-stage method SCG\cite{yan2018spatial} by 0.42\% map. However, the two stage pipeline includes heuristic processes such as NMS and is not end-to-end.

\begin{table*}[t]
    \centering
    \scalebox{0.9}{
    \begin{tabular}{c|cc|ccc|ccc}
      \hline
      &  &  &  \multicolumn{3}{|c|}{Default} & \multicolumn{3}{|c}{Known Object} \\
      Method & Detector & Backbone & Full & Rare  & Non-Rare & Full & Rare & Non-rare \\
      \hline
      Two-stage Method & & & & & & & & \\
      GPNN\cite{qi2018learning} & COCO & R101 & 13.11 & 9.34 & 14.23 & - & - & - \\
      iCAN\cite{gao2018ican} & COCO & R50 & 14.84 & 10.45 & 16.15 & 16.26 & 11.33 & 17.73 \\
      DCA\cite{wang2019deep} & COCO & R50 & 16.24 & 11.16 & 17.75 & 17.73 & 12.78 & 19.21 \\
      TIN\cite{li2019transferable} & COCO & R50 & 17.03 & 13.42 & 18.11 & 19.17 & 15.51 & 20.26 \\
      RPNN\cite{zhou2019relation} & COCO & R50 & 17.35 & 12.78 & 18.71 & - & - & -  \\
      PMFNet\cite{wan2019pose} & COCO & R50-FPN & 17.46 & 15.65 & 18.00 & 20.34 & 17.47 & 21.20 \\
      FCMNet\cite{liu2020amplifying} & COCO & R50 & 20.41 & 17.34 & 21.56 & 22.04 & 18.97 & 23.12 \\
      DJ-RN\cite{li2020detailed} & COCO & R50 & 21.34 & 18.53 & 22.18 & 23.69 & 20.64 & 24.60 \\
      IDN\cite{li2020hoi} & COCO & R50 & 23.36 & 22.47 & 23.63 & 26.43 & 25.01 & 26.85 \\
      VCL\cite{hou2020visual} & HICO-DET & R50 & 23.63 & 17.21 & 25.55 & 25.98 & 19.12 & 28.03 \\
      DRG\cite{gao2020drg} & HICO-DET & R50-FPN & 24.53 & 19.47 & 26.04 & 27.98 & 23.11 & 29.43 \\
      IDN\cite{li2020hoi} & HICO-DET & R50 & 24.58 & 20.33 & 25.86 & 27.89 & 23.64 & 29.16 \\
      SCG\cite{zhang2021spatially} & HICO-DET & R50-FPN & 31.33 & 24.72 & 33.31 & 34.37 & 27.18 & 36.52 \\ 
      \hline
      One-stage Method & & & & & \\
      UnionDet\cite{kim2020uniondet} & HICO-DET & R50-FPN & 17.58 & 11.72 & 19.33 & 19.76 & 14.68 & 21.27 \\
      IPNet\cite{wang2020learning} & COCO & R50-FPN &  19.56 & 12.79 & 21.58 & 22.05 & 15.77 & 23.92 \\
      PPDM\cite{liao2020ppdm} & HICO-DET & HG104 & 21.94 & 13.97 & 24.32 & 24.81 & 17.09 & 27.12 \\
      DIRV\cite{fang2020dirv} & HICO-DET & EfficientDet-d3 & 21.78 & 16.38 & 23.39 & 25.52 & 20.84 & 26.92 \\
      HOTR\cite{hotr} & HICO-DET & R50 &  25.10 & 17.34 & 27.42 & - & - & - \\
      HOITrans\cite{hoitrans} & HICO-DET & R101 & 26.61 & 19.15 & 28.84 & 29.13 & 20.98 & 31.57 \\
      AS-Net\cite{asnet} & HICO-DET & R50 & 28.87 & 24.25 & 30.25 & 31.74 & 27.07 & 33.14 \\
      QPIC\cite{qpic} & HICO-DET & R50 & 29.07 & 21.85 & 31.23 & 31.68 & 24.14 & 33.93 \\
      QPIC\cite{qpic} & HICO-DET & R101 & 29.90 & 23.92 & 31.69 & 32.38 & 26.06 & 34.27 \\
      Ours & HICO-DET & R50 & \textbf{31.75} & \textbf{27.45} & \textbf{33.03} & \textbf{34.50} & \textbf{30.13} & \textbf{35.81} \\
      \hline
    \end{tabular}}
    \vspace{-0.15cm}
    \caption{\small Performance comparison on HICO-DET. 'COCO' means the object detector is freeze and pretrained on MS-COCO, 'HICO-DET' means the model is fine-tuned on HICO-DET training set.}
    \vspace{-0.2cm}
    \label{HICO-DET}
\end{table*}

\begin{table}[t]
    \centering
    \scalebox{0.9}{
    \begin{tabular}{c|cc|cc}
    \hline
    & base & head &  Scenario \#1 & Scenario \#2 \\
    \hline
    \multirow{3}{*}{Decoder}& 1 & 5 & 65.6 & 67.5 \\
    & 2 & 4 & \textbf{66.2} & \textbf{68.5} \\
    & 3 & 3 & 64.7 & 66.5 \\
    \hline
    \hline
    \multirow{3}{*}{Encoder} & 5 & 1 & 65.6 & 67.6 \\
    & 4 & 2 & \textbf{66.2} & \textbf{68.5} \\
    & 3 & 3 & 65.1 & 67.1\\
    \hline
    \end{tabular}}
    \vspace{-0.1cm}
    \caption{\small Ablation study on different transformer layers of base encoder/decoder and disentangled head encoders/decoders on VCOCO test set.}
    \vspace{-0.1cm}
    \label{layer ablation}
\end{table}


\subsection{Ablation Study} 
\paragraph{w/o encoder disentanglement}
Our model adopts a disentangled encoder to extract global contexts at three levels for different decoding sub-tasks. We replace the disentangled encoder with a single encoder of same layer in our full model, the performance drops 0.7\% mAP and 0.96\% mAP on both V-COCO and HICO-DET datasets respectively, as shown in Tab. \ref{module ablation}.

\vspace{-0.2cm}
\paragraph{w/o attentional fusion} Our attentional fusion block provides communications between two task decoders. As shown in Tab. \ref{module ablation}, we remove the attentional fusion block, the performance drops 1.8\% mAP and 0.51\% mAP on V-COCO and HICO-DET datasets respectively.
\vspace{-0.3cm}
\paragraph{w/o decoder disentanglement}
Our disentangled decoder is the key in our framework. It predicts interactive human-object instance pairs instead of individual objects as in prior parallel-branch transformers\cite{hotr,asnet}, and exploits unified HOI representation to associate instances and interactions. Without our decoder disentanglement, our model is more like the QPIC\cite{qpic}. Hence we compare the performances of ours and the single-branch transformer in Tab. \ref{module ablation}. We can observe that performances significantly drop on both datasets.

\begin{table}[t]
    \centering
    \scalebox{0.9}{
    \begin{tabular}{c|cc}
    \hline
        Method & VCOCO & HICO \\
        \hline
        feature decomposition(proposed) & \textbf{66.2} & \textbf{31.75} \\
        query decomposition & 64.9 & 31.09 \\
        \hline
    \end{tabular}}
    \vspace{-0.1cm}
    \caption{\small Different association strategies of instances and interactions on V-COCO test set (Scenario \#1) and HICO-DET test set (Default, Full setting)}
    \vspace{-0.2cm}
    \label{association}
\end{table}



\begin{table}[t]
    \centering
    \scalebox{0.9}{
    \begin{tabular}{c|cc}
    \hline
        Method & VCOCO & HICO \\
        \hline
        w/o warmup & 65.7 & 31.49 \\
        w/ warmup &  \textbf{66.2} & \textbf{31.75} \\
        \hline
    \end{tabular}}
    \vspace{-0.2cm}
    \caption{\small Effect of warmup strategy on V-COCO test set (Scenario \#1) and HICO-DET test set (Default, Full setting)}
    \vspace{-0.3cm}
    \label{warmup}
\end{table}

\vspace{-0.3cm}
\paragraph{Effect of warmup strategy}
Since our transformer model has more parameters than original DETR, we adopt a warmup strategy during training. To validate the effectiveness of our warmup strategy, we perform an ablation study about the warmup strategy, shown in Tab.\ref{warmup}. We notice that the warmup strategy slightly improves the performances on both datasets.

\vspace{-0.3cm}
\paragraph{Different layers of base/head encoders/decoders}
We further perform ablation study on different transformer layers of base encoder/decoder and disentangled head encoders/decoders, shown in Tab.\ref{layer ablation}. For simplicity of our model and usage of pre-trained DETR parameters, we empirically keep the sum of base layer and head layer to 6, as in the original transformer. From the first three rows, we can observe that the decoder base layer $L_{de}^b=2$ and head layer $L_{de}^h=4$ is the best proportion and provides best performance, demonstrating the importance of unified representation. From the bottom three rows, we can see that 4-layer base with 2-layer head outperforms 3-layer base with 3-layer head, which implies that the modeling of shared global contexts in base encoder is also important.



\begin{figure*}[t]
    \includegraphics[width=\textwidth]{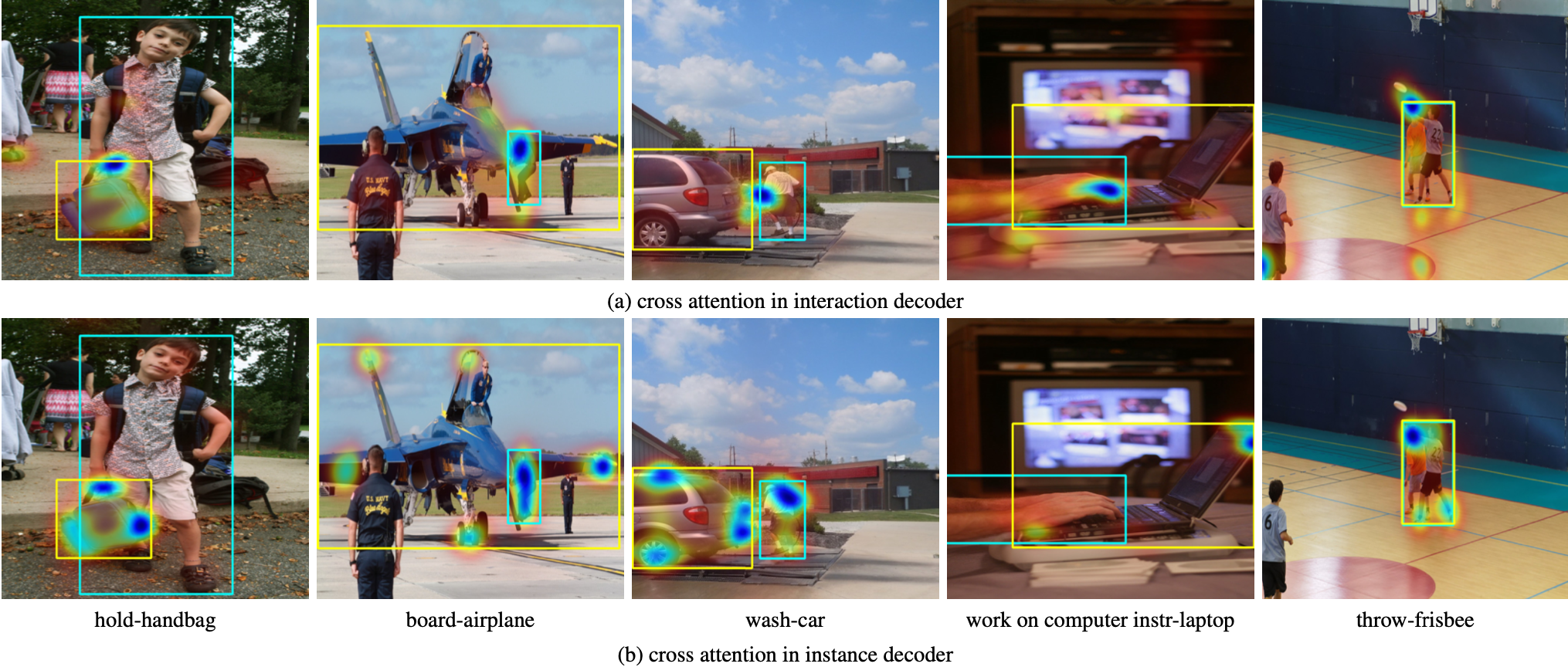}
    \vspace{-0.3cm}
    \caption[]{Visualization of cross attention maps of the same triplet prediction in our interaction decoder(top row) and instance decoder(bottom row). The left three samples are from HICO-DET\cite{hico} and others are from V-COCO\cite{vcoco}. In the top row, we can see that our interaction decoder attends to the interactive regions of human and objects. In the bottom row, we can see that our instance decoder attends to the object extremities. The different regions the model attends to implies that interaction and instance decoders indeed capture the disentangled representations of images.}
    \vspace{-0.6cm}
    \label{attention}
\end{figure*}

\vspace{-0.3cm}

\paragraph{Different association strategies}
Different from previous parallel-branch HOI transformer\cite{asnet,qpic} that instance decoder predicts individual objects in the image, our instance decoder directly estimates a set of interactive human-object instance pairs. In our framework, we adopt a base decoder to generate a unified representation to associate the estimated human-object instance pairs and interactions. We notice that there might be different association strategies. To study the effectiveness of our coarse-to-fine association strategy(referred to as feature decomposition), we replace the unified representation with a set of learnable unified HOI queries, which are then used to generate two queries with MLPs for disentangled decoders(referred to as query decomposition). We keep our disentangled encoder and attentional fusion block for fair comparison. As shown in Tab. \ref{association}, the performance drops by 1.3\% mAP and 0.66\% mAP on V-COCO and HICO-DET datasets respectively, which implies our association strategy is effective.

\begin{table}[t]
    \centering
    \scalebox{0.9}{
    \begin{tabular}{c|c|c|c|c}
    \hline
        Method & Backbone & AP & Params(M) & FLOPs(G) \\
        \hline
        QPIC\cite{qpic} & R50 & 58.8 & 41.68  & 87.87 \\
        QPIC\cite{qpic} & R101 & 58.3 & 60.62 & 156.18 \\
        AS-Net\cite{asnet}  & R50 & 53.9 & 52.75 & 88.86 \\
        HOTR\cite{hotr} & R50 & 55.2 & 51.41 & 88.78 \\
        HOITrans\cite{hoitrans} & R101 & 52.9 & 60.62 & 156 \\
        Ours & R50 & \textbf{66.2} & \textbf{57.31} & \textbf{94.23} \\
        \hline
    \end{tabular}}
    \vspace{-0.2cm}
    \caption{\small Model complexity comparison between ours and prior state-of-the-art HOI transformers. ‘AP’ indicates the performances on V-COCO test set under scenario \#1.}
    \vspace{-0.3cm}
    \label{tab:complexity}
\end{table}
\subsection{Model Complexity Analysis}
Since our model includes more encoder/decoder and fusion blocks, readers may care about the complexity of our model. Therefore, we compare the parameters and FLOPS of our final model and prior HOI Transformers in Tab .\ref{tab:complexity}. Similar to DETR\cite{detr}, we compute the FLOPS with the tool \textbf{flop\_count\_operators} from Detectron2\cite{wu2019detectron2} for the ﬁrst 100 images in the V-COCO test set and calculate the average numbers. We observe that our model has comparable parameters and FLOPS compared with prior HOI transformers. In particular, our model merely introduces 7\% extra FLOPS compared with the single-branch QPIC under R50, demonstrating both efficiency and effectiveness of our disentangled transformer.

\subsection{Qualitative Analysis}
As shown in Fig \ref{attention}, we visualize the cross attention maps of the same triplet prediction in instance decoder and interaction decoder. Top row shows the attention maps of interaction decoder, we can observe that the attention maps highlight the interactive regions between human-object instance pairs. In the bottom row, we can observe that the instance attention map attends to the object extremities, which is similar to DETR\cite{detr}. The different attention maps implies that our instance and interaction decoders indeed capture disentangled representations.

\vspace{-0.1cm}
\section{Conclusion}



In this paper, we propose disentangled transformer for HOI detection. Our method
decouples the triplet prediction into human-object pair detection and interaction classification via an instance stream and an interaction stream, where both encoder and decoder are disentangled. To associate the predictions of two task decoders, we adopt a coarse-to-fine strategy that first utilizes a base decoder to generate a unified HOI representation, and then conduct feature refinement in the disentangled instance and interaction spaces. We further propose an attentional fusion block to help two task decoders communicate with each other. 
As a result, our method is able to outperform prior HOI transformers and other methods by a sizeable margin on both V-COCO and HICO-DET benchmarks. The visualization of cross attention maps in task decoders also provide a good interpretation of the disentangled strategy.


\vspace{-0.1cm}
\section*{Potential Negative Societal Impact}
Our algorithm has no evident threats to society. However, someone might use our method for malicious usage, e.g. to attack people in military usage or invasion of privacy with surveillance. Therefore, we encourage good faith consideration before adopting our technology.


\newpage
{\small
\bibliographystyle{ieee_fullname}
\bibliography{egbib}
}

\end{document}